# Identifying Conditional Causal Effects

**Jin Tian**
Department of Computer Science
Iowa State University
Ames, IA 50011
*jtian@cs.iastate.edu*

## Abstract

This paper concerns the assessment of the effects of actions from a combination of nonexperimental data and causal assumptions encoded in the form of a directed acyclic graph in which some variables are presumed to be unobserved. We provide a procedure that systematically identifies cause effects between two sets of variables conditioned on some other variables, in time polynomial in the number of variables in the graph. The identifiable conditional causal effects are expressed in terms of the observed joint distribution.

## 1 Introduction

This paper explores the feasibility of inferring cause effect relationships from various combinations of data and theoretical assumptions. The assumptions considered will be represented in the form of an acyclic causal graph containing unmeasured variables [Pearl, 1995, Pearl, 2000] in which arrows represent the potential existence of direct causal relationships between the corresponding variables. Our main task will be to decide whether the assumptions represented in any given graph are sufficient for assessing the strength of causal effects from nonexperimental data and, if sufficiency is proven, to express the target causal effect in terms of estimable quantities.

It is well known that, in the absence of unmeasured variables, all causal effects are *identifiable*, that is, the joint response of any set $S$ of variables to intervention on a set $T$ of action variables, denoted $P_t(s)$,[1] can be estimated consistently from nonexperimental data [Robins, 1986, Spirtes *et al.*, 1993, Pearl, 1993]. If some variables are not measured, then the question of identifiability arises, and whether the desired quantity can be estimated depends critically on the precise locations (in the graph) of those unmeasured variables vis a vis the sets $T$ and $S$. Sufficient graphical conditions for ensuring the identification of $P_t(s)$ were established by several authors, for example, the popular "back-door" and "front-door" criteria [Pearl, 1995]. More generally, identifiability can be decided using Pearl's *do*-calculus [Pearl, 1995], which includes three inference rules by which probabilistic sentences involving interventions and observations can be transformed into other such sentences. A finite sequence of syntactic transformations, each applying one of the inference rules, may reduce expressions of the type $P_t(s)$ to subscript-free expressions involving observed quantities. Using *do*-calculus as a guide, [Galles and Pearl, 1995] devised a graphical criterion for identifying $P_x(y)$ where $X$ and $Y$ are singletons that combines and expands the "front-door" and "back-door" criteria.[2] [Pearl and Robins, 1995] derived a graphical condition under which it is possible to identify $P_t(y)$ where $T$ consists of an arbitrary set of variables and $Y$ is a singleton. This permits one to evaluate the effects of (unconditional) plans in the presence of unmeasured variables, each plan consisting of several concurrent or sequential actions. This criterion was further extended by [Robins, 1997] and [Kuroki and Miyakawa, 1999]. These results are summarized in [Pearl, 2000, Chapters 3 and 4].

The main difficulty with applying *do*-calculus for identifying causal effects is that there is no general guidance on which *do*-calculus rule to apply at each step so as to finally decide whether a causal effect is identifiable or not. Although many sufficient graphical criteria have been derived, they are applicable only when required graphical conditions are satisfied. Recently, a new approach for identifying causal effects was developed in [Tian and Pearl, 2002, Tian and Pearl, 2003], based on a factorization of the observed distribution in the presence of unmeasured variables. A systematic procedure was pro-vided for identifying the causal effects $P_t(s)$ of a set of variables $T$ on another set $S$ [Tian and Pearl, 2003].

---

[1][Pearl, 1995, Pearl, 2000] used the notation $P(s|set(t))$, $P(s|do(t))$, or $P(s|\hat{t})$ for the post-intervention distribution, while [Lauritzen, 2000] used $P(s||t)$.

[2][Galles and Pearl, 1995] claimed that their criterion embraces all cases where identification is verifiable by *do*-calculus. But [Tian and Pearl, 2003] shows that their criterion is *not* complete in this sense.

In this paper, we study the problem of identifying $P_t(s|c)$, the causal effects of $T$ on $S$ conditioned on another set $C$ of variables. This problem is important for evaluating *conditional plans* and stochastic plans [Pearl and Robins, 1995], where action $T$ is taken to respond in a specified way to a set $C$ of other variables – say, through a functional relationship $t = g(c)$ or through a stochastic relationship whereby $T$ is set to $t$ with some probability $P^*(t|c)$. It is shown in [Pearl, 2000, chapter 4] that the effects of such actions can be evaluated through identifying conditional causal effects in the form of $P_t(s|c)$. Evaluation of conditional plans has been considered in [Robins, 1986, Robins, 1987], and in [Kuroki *et al.*, 2003] in linear structural equation models.

We can certainly apply *do*-calculus for identifying conditional causal effects. Again, the difficulty lies in that there is no general heuristics as to how to use those inference rules. In this paper, we use the approach in [Tian and Pearl, 2003] and develop a procedure that systematically identifies conditional causal effects. The analysis of the paper will rely heavily on results reported in [Tian and Pearl, 2003] which is only concerned with identifying unconditional causal effects.

The rest of the paper is organized as follows. Section 2 introduces causal models and gives formal definition for the identifiability problem. Section 3 reviews the approach for identifying causal effects developed in [Tian and Pearl, 2003], and Section 4 shows how to identify conditional causal effects using this approach. Section 5 concludes the paper with discussions on future research.

## 2 Notation, Definitions, and Problem Formulation

The use of graphical models for encoding distributional and causal assumptions is now fairly standard [Spirtes *et al.*, 1993, Heckerman and Shachter, 1995, Lauritzen, 2000, Pearl, 2000]. The most common such representation involves a *Markovian model* (also known as a *causal Bayesian network*). A Markovian model consists of a DAG $G$ over a set $V = \{V_1, \ldots, V_n\}$ of variables, called a *causal graph*. The interpretation of such a graph has two components, probabilistic and causal. The probabilistic interpretation views $G$ as representing conditional independence assertions: Each variable is independent of all its non-descendants given its direct parents in the graph. These assertions imply that the joint probability function $P(v) = P(v_1, \ldots, v_n)$ factorizes according to the product [Pearl, 1988]

$$P(v) = \prod_i P(v_i|pa_i) \qquad (1)$$

where $pa_i$ are (values of) the parents of variable $V_i$ in the graph.[3]

The causal interpretation views the directed edges in $G$ as representing causal influences between the corresponding variables. This additional assumption enables us to predict the effects of interventions, whenever interventions are described as specific modifications of some factors in the product of (1). The simplest such intervention involves fixing a set $T$ of variables to some constants $T = t$, denoted by the action $do(T = t)$ or simply $do(t)$, which yields the post-intervention distribution

$$P_t(v) = \begin{cases} \prod_{\{i|V_i \notin T\}} P(v_i|pa_i) & v \text{ consistent with } t. \\ 0 & v \text{ inconsistent with } t. \end{cases} \qquad (2)$$

Eq. (2) represents a truncated factorization of (1), with factors corresponding to the manipulated variables removed. If $T$ stands for a set of treatment variables and $Y$ for an outcome variable in $V \setminus T$, then Eq. (2) permits us to calculate the probability $P_t(y)$ that event $Y = y$ would occur if treatment condition $T = t$ were enforced uniformly over the population. This quantity, often called the *causal effect* of $T$ on $Y$, is what we normally assess in a controlled experiment with $T$ randomized, in which the distribution of $Y$ is estimated for each level $t$ of $T$.

We see that, whenever all variables in $V$ are observed, given the causal graph $G$, all causal effects can be computed from the observed distribution $P(v)$. Our ability to estimate $P_t(v)$ from observed data is severely curtailed when some variables are unobserved, or, equivalently, if two or more variables in $V$ are affected by unobserved confounders; the presence of such confounders would not permit the decomposition of the observed distribution $P(v)$ in (1). Let $V$ and $U$ stand for the sets of observed and unobserved variables, respectively. In this paper, we assume that no $U$ variable is a descendant of any $V$ variable (called a *semi-Markovian* model). Then the observed probability distribution, $P(v)$, becomes a mixture of products:

$$P(v) = \sum_u \prod_i P(v_i|pa_i, u^i) P(u) \qquad (3)$$

where $Pa_i$ and $U^i$ stand for the sets of the observed and unobserved parents of $V_i$, and the summation ranges over all the $U$ variables. The post-intervention distribution, likewise, will be given as a mixture of truncated products

$$P_t(v) = \begin{cases} \sum_u \prod_{\{i|V_i \notin T\}} P(v_i|pa_i, u^i) P(u) & v \text{ consistent with } t. \\ 0 & v \text{ inconsistent with } t. \end{cases} \qquad (4)$$

---

[3]We use uppercase letters to represent variables or sets of variables, and use corresponding lowercase letters to represent their values (instantiations).

And, the question of identifiability arises, i.e., whether it is possible to express some $P_t(s)$ as a function of the observed distribution $P(v)$.

Formally, our semi-Markovian model consists of five elements

$$M = \langle V, U, G_{VU}, P(v_i|pa_i, u^i), P(u) \rangle$$

where $G_{VU}$ is a causal graph consisting of variables in $V \times U$. Clearly, given $M$ and any three sets $T$, $S$, and $C$ in $V$, $P_t(s|c)$ can be determined unambiguously using (4). The question of identifiability is whether a given (conditional) causal effect $P_t(s|c)$ can be determined uniquely from the distribution $P(v)$ of the observed variables, and is thus independent of the unknown quantities, $P(u)$ and $P(v_i|pa_i, u^i)$, that involve elements of $U$.

In order to analyze questions of identifiability, it is convenient to represent our modeling assumptions in the form of a graph $G$ that does not show the elements of $U$ explicitly but, instead, represents the confounding effects of $U$ using (dashed) bidirected edges. A bidirected edge between nodes $V_i$ and $V_j$ represents the presence of unobserved confounders that may influence both $V_i$ and $V_j$. See Figure 2(a) for an example graph with bidirected edges.

**Definition 1 (Causal-Effect Identifiability)** *The* causal effect *of a set of variables $T$ on a disjoint set of variables $S$ conditioned on another set $C$ is said to be* identifiable *from a graph $G$ if the quantity $P_t(s|c)$ can be computed uniquely from any positive probability of the observed variables—that is, if $P_t^{M_1}(s|c) = P_t^{M_2}(s|c)$ for every pair of models $M_1$ and $M_2$ with $P^{M_1}(v) = P^{M_2}(v) > 0$ and $G(M_1) = G(M_2) = G$.*

In other words, the quantity $P_t(s|c)$ can be determined from the observed distribution $P(v)$ alone; the details of $M$ are irrelevant.

## 3 Q-decomposition and Causal Effects Identification

In this section, we review the techniques developed in [Tian and Pearl, 2003] for identifying causal effects.

Let a path composed entirely of bidirected edges be called a *bidirected path*. The set of variables $V$ in $G$ can be partitioned into disjoint groups by assigning two variables to the same group if and only if they are connected by a bidirected path. Assuming that $V$ is thus partitioned into $k$ groups $S_1, \ldots, S_k$, each set $S_j$ is called a *c-component* of $V$ in $G$ or a c-component of $G$. For example, the graph in Figure 2(a) consists of three c-components: $\{B\}$, $\{Z\}$, and $\{A, X, W, Y\}$.

For any set $C \subseteq V$, define the quantity $Q[C](v)$ to denote the post-intervention distribution of $C$ under an intervention to all other variables:[4]

$$Q[C](v) = P_{v \setminus c}(c) = \sum_u \prod_{\{i|V_i \in C\}} P(v_i|pa_i, u^i)P(u). \quad (5)$$

In particular, we have $Q[V](v) = P(v)$. If there is no bidirected edges connected with a variable $V_i$, then $U^i = \emptyset$ and $Q[\{V_i\}] = P(v_i|pa_i)$. Let $Pa(S)$ denote the union of a set $S$ and the set of parents of $S$, that is, $Pa(S) = S \cup (\cup_{V_i \in S} Pa_i)$. From Eq. (5) we see that $Q[C]$ is actually a function of $Pa(C)$. For convenience, we will often write $Q[C](pa(C))$ as $Q[C]$.

The importance of the c-component comes from the following lemma.

**Lemma 1 (Q-decomposition)** *[Tian and Pearl, 2002] Assuming that $V$ is partitioned into c-components $S_1, \ldots, S_k$, we have*

*(i)* $P(v) = \prod_i Q[S_i]$.

*(ii) Each $Q[S_i]$ is computable from $P(v)$. Let a topological order over $V$ be $V_1 < \ldots < V_n$, and let $V^{(i)} = \{V_1, \ldots, V_i\}$, $i = 1, \ldots, n$, and $V^{(0)} = \emptyset$. Then each $Q[S_j]$, $j = 1, \ldots, k$, is given by*

$$Q[S_j] = \prod_{\{i|V_i \in S_j\}} P(v_i|v^{(i-1)}) \quad (6)$$

The lemma says that $P(v)$ can be decomposed into a product of $Q[S_i]$'s (called *Q-decomposition*) and each $Q[S_i] = P_{v \setminus s_i}(s_i)$ is identifiable. For example, in Figure 2(a), letting $S_1 = \{A, X, W, Y\}$, we have

$$\begin{aligned} P(v) &= Q[\{B\}]Q[\{Z\}]Q[S_1] \\ &= P(b|a)P(z|x)Q[S_1], \end{aligned} \quad (7)$$

in which $Q[S_1] = P_{bz}(a, x, w, y)$ is given by

$$Q[S_1] = P(y|z, w, x, b, a)P(w|x, b, a)P(x|b, a)P(a) \quad (8)$$

For any set $C$, let $G_C$ denote the subgraph of $G$ composed only of variables in $C$. Lemma 1 can be generalized to the subgraphs of $G$ as given in the following lemma.

**Lemma 2 (Generalized Q-decomposition)**
*[Tian and Pearl, 2003] Let $H \subseteq V$, and assume that $H$ is partitioned into c-components $H_1, \ldots, H_l$ in the subgraph $G_H$. Then we have*

*(i) $Q[H]$ decomposes as*

$$Q[H] = \prod_i Q[H_i]. \quad (9)$$

*(ii) Each $Q[H_i]$ is computable from $Q[H]$. Let $k$ be the number of variables in $H$, and let a topological order of*

---

[4]Set $Q[\emptyset](v) = 1$ since $\sum_u P(u) = 1$.

*the variables in H be $V_{h_1} < \cdots < V_{h_k}$ in $G_H$. Let $H^{(i)} = \{V_{h_1}, \ldots, V_{h_i}\}$ be the set of variables in H ordered before $V_{h_i}$ (including $V_{h_i}$), $i = 1, \ldots, k$, and $H^{(0)} = \emptyset$. Then each $Q[H_j]$, $j = 1, \ldots, l$, is given by*

$$Q[H_j] = \prod_{\{i|V_{h_i} \in H_j\}} \frac{Q[H^{(i)}]}{Q[H^{(i-1)}]}, \quad (10)$$

*where each $Q[H^{(i)}]$, $i = 0, 1, \ldots, k$, is given by*

$$Q[H^{(i)}] = \sum_{h \setminus h^{(i)}} Q[H]. \quad (11)$$

Lemma 2 says that if $Q[H] = P_{v \setminus h}(h)$ is identifiable and the subgraph $G_H$ is partitioned into c-components $H_1, \ldots, H_l$, then each $Q[H_i] = P_{v \setminus h_i}(h_i)$ is identifiable.

For any set $S$, let $An(S)$ denote the union of $S$ and the set of ancestors of the variables in $S$. For $W \subseteq C \subseteq V$, the following lemma gives a condition under which $Q[W]$ can be computed from $Q[C]$ by summing over $C \setminus W$, like ordinary marginalization in probability theory.

**Lemma 3** *[Tian and Pearl, 2003] Let $W \subseteq C \subseteq V$, and $W' = C \setminus W$. If W contains its own ancestors in the subgraph $G_C$ ($An(W)_{G_C} = W$), then*

$$\sum_{w'} Q[C] = Q[W]. \quad (12)$$

Note that we always have $\sum_c Q[C] = 1$.

Next, we show how to use Lemmas 1–3 to identify the causal effect $P_t(s)$ where $S$ and $T$ are arbitrary (disjoint) subsets of $V$. We have

$$P_t(s) = \sum_{(v \setminus t) \setminus s} P_t(v \setminus t) = \sum_{(v \setminus t) \setminus s} Q[V \setminus T]. \quad (13)$$

Let $D = An(S)_{G_{V \setminus T}}$. Then by Lemma 3, variables in $(V \setminus T) \setminus D$ can be summed out:

$$P_t(s) = \sum_{d \setminus s} \sum_{(v \setminus t) \setminus d} Q[V \setminus T] = \sum_{d \setminus s} Q[D]. \quad (14)$$

Assume that the subgraph $G_D$ is partitioned into c-components $D_1, \ldots, D_l$. Then by Lemma 2, $Q[D]$ can be decomposed into products of $Q[D_i]$'s, and Eq. (14) can be rewritten as

$$P_t(s) = \sum_{d \setminus s} \prod_i Q[D_i]. \quad (15)$$

We obtain that $P_t(s)$ is identifiable if all $Q[D_i]$'s are identifiable.

Let $G$ be partitioned into c-components $S_1, \ldots, S_k$. Then any $D_i$ is a subset of certain $S_j$ since if the variables in

**Algorithm Identify**$(C, T, Q)$
INPUT: $C \subseteq T \subseteq V$, $Q = Q[T]$. Assuming $G_T$ is composed of one single c-component.
OUTPUT: Expression for $Q[C]$ in terms of $Q$ or FAIL.

Let $A = An(C)_{G_T}$.

- IF $A = C$, output $Q[C] = \sum_{t \setminus c} Q$.
- IF $A = T$, output FAIL.
- IF $C \subset A \subset T$
  1. Assume that in $G_A$, $C$ is contained in a c-component $T'$.
  2. Compute $Q[T']$ from $Q[A] = \sum_{t \setminus a} Q$ by Lemma 2.
  3. Output Identify$(C, T', Q[T'])$.

Figure 1: An algorithm for determining if $Q[C]$ is computable from $Q[T]$ [Tian and Pearl, 2003].

$D_i$ are connected by a bidirected path in a subgraph of $G$ then they must be connected by a bidirected path in $G$. Assuming $D_i \subseteq S_j$, [Tian and Pearl, 2003] gives a recursive algorithm Identify$(D_i, S_j, Q[S_j])$, shown in Figure 1, for determining whether $Q[D_i]$ is computable from $Q[S_j]$ which is identified by Lemma 1. The algorithm works by repeated applications of Lemma 3 and 2, and outputs either an expression for $Q[D_i]$ in terms of $Q[S_j]$ or FAIL.

The algorithm in Figure 1 fails to compute $Q[C]$ from $Q[T]$ if all of the following three conditions are satisfied: (i) $G_C$ has only one c-component ($C$ itself), (ii) $G_T$ has only one c-component ($T$ itself), and (iii) all variables in $T \setminus C$ are ancestors of $C$ in $G_T$ ($An(C)_{G_T} = T$). Whether $Q[C]$ is indeed unidentifiable in this situation is still an open problem.

To summarize, the identifiability of causal effect $P_t(s)$ is reduced to the identifiability of some $Q[D_i]$'s, which is determined by using the algorithm Identify$(\cdot, \cdot, \cdot)$. In the rest of the paper, when we refer to "unidentifiable" we will mean unidentifiable through the use of algorithm Identify$(\cdot, \cdot, \cdot)$.

## 4 Identifying Conditional Causal Effects

In this section we study the problem of identifying conditional causal effects $P_t(s|c)$ where $S$, $T$, and $C$ are (disjoint) subsets of $V$. We have

$$P_t(s|c) = \frac{P_t(s, c)}{P_t(c)}. \quad (16)$$

Therefore, first of all, $P_t(s|c)$ is identifiable if $P_t(s, c)$ is identifiable. We can use the procedure in [Tian and Pearl, 2003] illustrated in Section 3 to identify $P_t(s,c)$ and, if the answer is positive, then $P_t(s|c)$ is also identified. Sufficient graphical criteria developed for identifying (unconditional) causal effects can also be used as sufficient criteria for conditional causal effects. In particular, there exists a powerful graphical criterion for the causal effects of a singleton variable, given in the following.

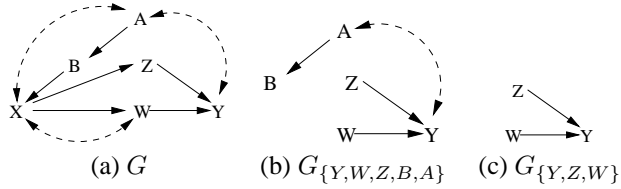

Figure 2: A causal graph

**Theorem 1** *[Tian and Pearl, 2002] Let $X$ be a singleton variable and let $S$ and $C$ be two sets of variables. $P_x(s|c)$ is identifiable if there is no bidirected path connecting $X$ to any of its children in the subgraph $G_{An(S \cup C)}$.*

This theorem, although not complete, is useful for human experts to make quick judgement of identifiability by looking at the causal graph. For example, by Theorem 1, in the graph Figure 2(a), the conditional causal effect $P_a(s|c)$ is identifiable for any sets $S, C \subseteq \{B, X, W, Z, Y\}$.

Secondly, if $P_t(s,c)$ is not identifiable but $P_t(c)$ is, then $P_t(s|c)$ is not identifiable. A special case is if none of the variables in $C$ is a descendant of variables in $T$, then $P_t(c) = P(c)$, and $P_t(s|c)$ is identifiable if and only if $P_t(s,c)$ is identifiable. This condition is often satisfied in conditional plan problems in which the set of conditional variables $C$ is normally assumed not to contain variables influenced by the control variables $T$ [Pearl and Robins, 1995].

Finally, the most interesting case is when neither $P_t(s,c)$ nor $P_t(c)$ is identifiable. Yet, $P_t(s|c)$ may still be identifiable in this case. Next, we work out an example in which both $P_t(s,c)$ and $P_t(c)$ is unidentifiable but $P_t(s|c)$ is identifiable, and the computing process will give us hints on the conditions for $P_t(s|c)$ to be identifiable.

Consider the problem of identifying $P_x(y|w)$ in Figure 2(a). First we compute $P_x(y,w)$ as

$$P_x(y,w) = \sum_{z,b,a} P_x(y,w,z,b,a)$$

$$= \sum_{z,b,a} Q[\{Y,W,Z,B,A\}]$$

$$= \sum_z Q[\{Y,Z,W\}] \quad \text{(Lemma 3 in Figure 2(b))}$$

$$= \sum_z Q[\{Z\}]Q[\{W\}]Q[\{Y\}] \quad (17)$$

(Lemma 2 in Figure 2(c))

The graph $G$ in Figure 2(a) is partitioned into c-components $\{B\}$, $\{Z\}$, and $S_1 = \{A, X, W, Y\}$ with $Q[\{Z\}] = P(z|x)$ and $Q[S_1]$ given in Eq. (8). Calling the algorithm Identify($\{W\}, S_1, Q[S_1]$) in Figure 1, it turns out that $Q[\{W\}]$ is unidentifiable. Calling the function Identify($\{Y\}, S_1, Q[S_1]$), we obtain that $Q[\{Y\}]$ is identifiable and is given by

$$Q[\{Y\}] = \frac{\sum_a Q[S_1]}{\sum_{a,y} Q[S_1]}. \quad (18)$$

Eq. (17) can be rewritten as

$$P_x(y,w) = Q[\{W\}](w,x) \sum_z P(z|x)Q[\{Y\}], \quad (19)$$

in which $Q[\{Y\}]$ is identifiable and $Q[\{W\}]$ is not. We conclude that $P_x(y,w)$ is unidentifiable. $P_x(w)$ can be computed from $P_x(y,w)$ as

$$P_x(w) = \sum_y P_x(y,w) = Q[\{W\}] \quad (20)$$

$$(\sum_y Q[\{Y\}] = 1)$$

Therefore $P_x(w)$ is not identifiable either. From Eq. (19) and (20), we obtain

$$P_x(y|w) = \frac{P_x(y,w)}{P_x(w)} = \sum_z P(z|x)Q[\{Y\}], \quad (21)$$

which is identifiable. The reason for $P_x(y|w)$ being identifiable is that the unidentifiable term $Q[\{W\}]$ is canceled out which appears in both $P_x(y,w)$ and $P_x(w)$.

Now if we would like to identify $P_x(y|w,b)$ in Figure 2(a), first we compute $P_x(y,w,b)$ as

$$P_x(y,w,b) = \sum_{z,a} Q[\{Y,W,Z,B,A\}]$$

$$= \sum_{a,z} Q[\{B\}]Q[\{Z\}]Q[\{W\}]Q[\{A,Y\}]$$

(Lemma 2 in Figure 2(b))

$$= Q[\{W\}](w,x) \sum_{a,z} P(b|a)P(z|x)Q[\{A,Y\}] \quad (22)$$

Calling the algorithm Identify($\{A, Y\}, S_1, Q[S_1]$) returns that $Q[\{A, Y\}]$ is unidentifiable. Therefore $P_x(y, w, b)$ is unidentifiable. $P_x(w, b)$ can be computed from $P_x(y, w, b)$ as

$$P_x(w, b) = \sum_y P_x(y, w, b)$$
$$= Q[\{W\}](w, x) \sum_y \sum_{a,z} P(b|a)P(z|x)Q[\{A, Y\}] \quad (23)$$

which is not identifiable either. From Eq. (22) and (23), we obtain

$$P_x(y|w, b) = \frac{\sum_{a,z} P(b|a)P(z|x)Q[\{A, Y\}]}{\sum_y \sum_{a,z} P(b|a)P(z|x)Q[\{A, Y\}]}, \quad (24)$$

which is still unidentifiable. $P_x(y|w, b)$ is unidentifiable because $Y$ appears in an unidentifiable term in $P_x(y, w, b)$, and therefore this term can not be canceled out by any term in $P_x(w, b)$.

The previous example shows that in the case when neither $P_t(s, c)$ nor $P_t(c)$ is identifiable, $P_t(s|c)$ may still be identifiable if the unidentifiable terms are canceled out in the expressions for $P_t(s, c)$ and $P_t(c)$. Next, we study the conditions for this canceling out to happen.

First we compute an expression for $P_t(s, c)$ with the approach shown in Section 3. Assume that $V$ is partitioned into c-components $S_1, \ldots, S_k$. Let $D = An(S \cup C)_{G_{V \setminus T}}$, $F = D \setminus (S \cup C)$. Assume that the subgraph $G_D$ is partitioned into c-components $D_1, \ldots, D_l$. Then we have (see Eq. (15))

$$P_t(s, c) = \sum_f \prod_i Q[D_i]. \quad (25)$$

Each $D_i$ is a subset of certain $S_j$ and the identifiability of $Q[D_i]$ can be determined by calling the algorithm Identify($D_i, S_j, Q[S_j]$) where $Q[S_j]$ is identified by Lemma 1. Let $I$ be the set of $D_i$'s that are identifiable, and let $N$ be the set of $D_i$'s that are unidentifiable. Eq. (25) can be rewritten as

$$P_t(s, c) = \sum_f (\prod_{D_i \in N} Q[D_i])(\prod_{D_i \in I} Q[D_i]). \quad (26)$$

Using the fact that each $Q[D_i]$ is a function of $Pa(D_i)$, letting $H = \cup_{D_i \in N} Pa(D_i)$, then the unidentifiable term is a function of $H$. Define function $g(h) = \prod_{D_i \in N} Q[D_i]$. Eq. (26) becomes

$$P_t(s, c) = \sum_f g(h) \prod_{D_i \in I} Q[D_i]. \quad (27)$$

This summation over $F$ can (sometimes) be decomposed into a product of two summations: one over unidentifiable terms and the other over identifiable terms. Assume that the summation over $F$ can be decomposed as

$$P_t(s, c) = (\sum_{f_0} g(h) \prod_{D_i \in I_0} Q[D_i])(\sum_{f_1} \prod_{D_i \in I_1} Q[D_i]), \quad (28)$$

where $F$ is partitioned into two sets $F_0$ and $F_1$, and $I$ is partitioned into two sets $I_0$ and $I_1$, such that $F_0$ and $I_0$ contain as fewer elements as possible so as to make the decomposition possible. We will see in a moment how to do the partitions. The term $\sum_{f_0} g(h) \prod_{D_i \in I_0} Q[D_i]$ is unidentifiable and is a function of $H' = H \cup (\cup_{D_i \in I_0} Pa(D_i))$ while the term $\sum_{f_1} \prod_{D_i \in I_1} Q[D_i]$ is identifiable. Now since $P_t(c) = \sum_s P_t(s, c)$, if none of the variables in $S$ appears in the unidentifiable term, that is, if $S \cap H' = \emptyset$, then

$$P_t(c) = (\sum_{f_0} g(h) \prod_{D_i \in I_0} Q[D_i]) \sum_s \sum_{f_1} \prod_{D_i \in I_1} Q[D_i], \quad (29)$$

and $P_t(s|c)$ is identifiable as

$$P_t(s|c) = \frac{P_t(s, c)}{P_t(c)} = \frac{\sum_{f_1} \prod_{D_i \in I_1} Q[D_i]}{\sum_s \sum_{f_1} \prod_{D_i \in I_1} Q[D_i]}. \quad (30)$$

On the other hand, if $S \cap H' \neq \emptyset$, then the unidentifiable term $\sum_{f_0} g(h) \prod_{D_i \in I_0} Q[D_i]$ can not be totally canceled out by $\sum_s \sum_{f_0} g(h) \prod_{D_i \in I_0} Q[D_i]$, and $P_t(s|c)$ can not be identified.

How do we partition $F$ and $I$ such that Eq. (27) can be decomposed as Eq. (28) and $F_0$ and $I_0$ contain as fewer elements as possible? First of all those $F$ variables that are in $H$ should belong to $F_0$. Then for those $Q[D_i]$'s that contain variables in $F_0$ ($Pa(D_i) \cap F_0 \neq \emptyset$), $D_i$ have to be put in $I_0$. This may cause some new $F$ variables to be put in $F_0$, which in turn may cause more $D_i$'s to be put in $I_0$. We need to repeat this process until no new variables can be added to $F_0$. The above analysis leads to the following procedure for partitioning $F$ and $I$:

1. Initialize $F_0 = F \cap H$, $F_1 = F \setminus F_0$, and $I_0 = \emptyset$, $I_1 = I$.

2. For each $D_i \in I_1$, if $Pa(D_i) \cap F_0 \neq \emptyset$, then remove $D_i$ from $I_1$ and put it into $I_0$.

3. Let $B = F_1 \cap \cup_{D_i \in I_0} Pa(D_i)$.
   - If $B$ is not empty, remove variables in $B$ from $F_1$ and put them into $F_0$. Then go back to step 2.
   - If $B$ is empty, then stop. The partition process is finished.

In summary, an algorithm for computing $P_t(s|c)$ is given in Figure 3. The procedure consists of four basic phases. In phase-1, we compute the expressions for $Q[S_i]$'s and find

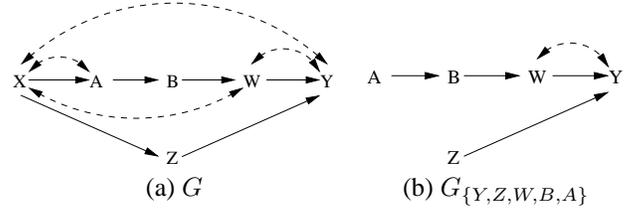

(a) $G$  (b) $G_{\{Y,Z,W,B,A\}}$

Figure 4: An example causal graph

**Algorithm 1 (Computing $P_t(s|c)$)**
*INPUT: three disjoint sets $T, S, C \subset V$.*
*OUTPUT: the expression for $P_t(s|c)$ or FAIL.*
*Phase-1:*

1. Find the c-components of $G$: $S_1, \ldots, S_k$.

2. Compute $Q[S_1], \ldots, Q[S_k]$ by Lemma 1.

3. Let $D = An(S \cup C)_{G_{V \setminus T}}$ and $F = D \setminus (S \cup C)$.

4. Find the c-components of the subgraph $G_D$: $D_1, \ldots, D_l$.

*Phase-2:*

1. For each set $D_i$:
   Find the set $S_j$ to which $D_i$ belongs. Call algorithm Identify$(D_i, S_j, Q[S_j])$. If the algorithm returns FAIL, then put $D_i$ into the set $N$, otherwise put $D_i$ into the set $I$.

2. If $N$ is empty, then stop and output

$$P_t(s|c) = \frac{\sum_f \prod_i Q[D_i]}{\sum_s \sum_f \prod_i Q[D_i]}$$

*Phase-3:*

1. Initialize $F_0 = F \cap (\cup_{D_i \in N} Pa(D_i))$, $F_1 = F \setminus F_0$, and $I_0 = \emptyset$, $I_1 = I$.

2. For each $D_i \in I_1$:
   If $Pa(D_i) \cap F_0 \neq \emptyset$, then remove $D_i$ from $I_1$ and put it into $I_0$.

3. Let $B = F_1 \cap \cup_{D_i \in I_0} Pa(D_i)$.
   - If $B$ is not empty, remove variables in $B$ from $F_1$ and put them into $F_0$. Then go back to step 2.
   - If $B$ is empty, then continue to Phase-4.

*Phase-4:*
If $S \cap (\cup_{D_i \in N \cup I_0} Pa(D_i)) = \emptyset$, then
  Output the expression for $P_t(s|c)$ as given in Eq. (30).
Else
  Output FAIL.

Figure 3: An algorithm for computing $P_t(s|c)$

the sets $D_i$'s and $F$ from the graph $G$. In phase-2, we determine the identifiability of each $Q[D_i]$ by calling the algorithm Identify$(\cdot, \cdot, \cdot)$ in Figure 1, and put $D_i$'s into two sets: $I$ if identifiable and $N$ if not. In phase-3, we partition $F$ into two sets $F_0$ and $F_1$, and $I$ into two sets $I_0$ and $I_1$. In phase-4, we determine the identifiability of $P_t(s|c)$, and when identifiable, output the expression for $P_t(s|c)$ as given in Eq. (30).

The algorithm runs in time polynomial in the number of variables in the graph. The main operations in phase 1 are finding c-components of graphs and finding ancestors of variables. The algorithm Identify$(\cdot, \cdot, \cdot)$ runs in polynomial time. The partition process in Phase 3 is clearly polynomial.

We demonstrate the use of the algorithm for computing $P_t(s|c)$ with an example. Consider the problem of identifying $P_x(y|a)$ in Figure 4(a).

Phase-1:
The c-components of $G$ are $\{B\}$, $\{Z\}$, and $S_1 = \{X, A, W, Y\}$. By Lemma 1,

$$Q[\{B\}] = P(b|a), \quad Q[\{Z\}] = P(z|x), \quad (31)$$

and

$$Q[S_1] = P(y|z, w, b, a, x)P(w|b, a, x)P(a|x)P(x). \quad (32)$$

The ancestors of $Y$ and $A$ in the subgraph with $X$ removed are $D = \{A, B, W, Z, Y\}$, and

$$F = D \setminus \{A, Y\} = \{B, W, Z\}. \quad (33)$$

The c-components of the subgraph $G_D$ are $\{A\}$, $\{B\}$, $\{Z\}$, and $\{W, Y\}$ (see Figure 4(b)).

Phase-2:
$Q[\{B\}]$ and $Q[\{Z\}]$ are identifiable and are given in Eq. (31). Calling algorithm Identify$(\{W, Y\}, S_1, Q[S_1])$ returns that $Q[\{W, Y\}]$ is identifiable and is given by

$$Q[\{W, Y\}] = \sum_{x,a} Q[S_1], \quad (34)$$

where $Q[S_1]$ is given in Eq. (32). Calling algorithm Identify$(\{A\}, S_1, Q[S_1])$ returns that $Q[\{A\}]$ is unidentifiable. Therefore, we set

$$N = \{\{A\}\}, \quad I = \{\{B\}, \{Z\}, \{W, Y\}\}. \quad (35)$$

Phase-3:
We partition $F$ and $I$. It turns out

$$F_0 = \emptyset, \quad F_1 = F = \{B, W, Z\}. \tag{36}$$

And

$$I_0 = \emptyset, \quad I_1 = I = \{\{B\}, \{Z\}, \{W, Y\}\}. \tag{37}$$

Phase-4:
Since $Y$ does not appear in $Pa(\{A\})$, we conclude that $P_x(y|a)$ is identifiable and is given by (see Eq. (30))

$$P_x(y|a) = \frac{\sum_{b,w,z} Q[\{B\}]Q[\{Z\}]Q[\{W,Y\}]}{\sum_{y,b,w,z} Q[\{B\}]Q[\{Z\}]Q[\{W,Y\}]}, \tag{38}$$

where expressions for $Q[\{B\}]$, $Q[\{Z\}]$ are given in Eq. (31), and that of $Q[\{W, Y\}]$ in Eq. (34).

## 5 Conclusion and Future Research

We present a procedure that systematically identifies conditional cause effects in time polynomial in the number of variables in the graph. When the concerned causal effects are identifiable, closed-form expressions are obtained in terms of the observed joint distribution. One application of the result is in evaluating conditional plans.

In light of results in [Tian and Pearl, 2003], we now have procedures for systematically identifying general (conditional and unconditional) cause effects in polynomial time. Some future research topics that we are investigating includes: Are our procedures complete in the sense that all identifiable causal effects can be derived using those procedures? Do our procedures identify all the causal effects that could be identified by $do$-calculus? Is every causal effect identifiable by our procedures also identifiable by $do$-calculus? The key to answering these questions may lie in the answer to the following questions: are those causal effects that the algorithm Identify$(\cdot, \cdot, \cdot)$ fails to identify indeed unidentifiable? or at least are unidentifiable by $do$-calculus?

**Acknowledgements**

This research was partly supported by NSF grant IIS-0347846.